\documentclass[10pt,journal,compsoc]{IEEEtran}
\usepackage{booktabs}
\usepackage{amssymb}

\usepackage[ruled]{algorithm2e}
\usepackage{graphicx}
\usepackage{caption}
\ifCLASSOPTIONcompsoc

\usepackage[nocompress]{cite}
\else
  \usepackage{cite}
\fi

\SetKwData{Module}{Module}
\begin{document}

\hbadness=2000000000
\vbadness=2000000000
\hfuzz=100pt

\title{MaskFi: Unsupervised Learning of WiFi and Vision Representations for Multimodal Human Activity Recognition}

\author{%
  \IEEEauthorblockN{%
    Jianfei~Yang,
    Shijie~Tang,
    Yuecong~Xu,
    Yunjiao~Zhou,
    and~Lihua~Xie~\IEEEmembership{Fellow,~IEEE},
  }%
  
   \thanks{S.Tang, Y.Xu, Y. Zhou, and L. Xie are with the School of Electrical and Electronics Engineering, Nanyang Technological University, Singapore (\{yang0478; stang023; xuyu0014; yunjiao001\}@e.ntu.edu.sg; elhxie@ntu.edu.sg).}
   
    \thanks{J. Yang is with the School of Mechanical and Aerospace Engineering, Nanyang Technological University, Singapore. J. Yang is the corresponding author (yang0478@e.ntu.edu.sg).}

    \thanks{J. Yang and S. Tang contribute equally to this work.}
}

\markboth{}%
{Tang \MakeLowercase{\textit{et al.}}: Unsupervised Learning of WiFi and Vision for Multimodal HAR}

\maketitle

\begin{abstract}
Human activity recognition (HAR) has been playing an increasingly important role in various domains such as healthcare, security monitoring, and metaverse gaming. Though numerous HAR methods based on computer vision have been developed to show prominent performance, they still suffer from poor robustness in adverse visual conditions in particular low illumination, which motivates WiFi-based HAR to serve as a good complementary modality. Existing solutions using WiFi and vision modalities rely on massive labeled data that are very cumbersome to collect. In this paper, we propose a novel unsupervised multimodal HAR solution, MaskFi, that leverages only unlabeled video and WiFi activity data for model training. We propose a new algorithm, masked WiFi-vision modeling (MI$^2$M), that enables the model to learn cross-modal and single-modal features by predicting the masked sections in representation learning. Benefiting from our unsupervised learning procedure, the network requires only a small amount of annotated data for finetuning and can adapt to the new environment with better performance. We conduct extensive experiments on two WiFi-vision datasets collected in-house, and our method achieves human activity recognition and human identification in terms of both robustness and accuracy.

\end{abstract}


\section{Introduction}\label{sec:introduction}

\IEEEPARstart{W}{ith} the advancement of deep learning and computer vision, human activity recognition (HAR) has achieved remarkable performance and enables many applications, such as human-computer interaction, security systems, metaverse, and gaming~\cite{chen2021deep}. A common scenario is in the field of smart homes, where automated adjustments of home temperature, lighting, and even music can be performed by recognizing human activities and predicting their intentions, enhancing the quality of life and comfort~\cite{bianchiIOt}. Furthermore, HAR can be applied to many interdisciplinary fields, such as healthcare~\cite{yang2018carefi,avci2010activity}.

Current HAR solutions can be classified into two categories: device-based and device-free methods. Device-based HAR requires users to wear specialized sensors such as inertial measurement unit (IMU)~\cite{avci2010activity,luo2021binarized,huang2022channel}. Though it achieves satisfactory performances on simple activity recognition such as gait counting, it only captures the information at one point of the human body and thus cannot model very complex activity. Moreover, it is inconvenient to use especially for the elderly. As for device-free HAR such as camera-based recognition methods~\cite{bux2017}, they have significant advantages such as high granularity and low cost, which are therefore widely deployed with the Closed Circuit Television (CCTV) systems. However, camera-based methods also have limitations. In adverse illumination environments, in particular dark environments, or when obstructions are present, camera-based HAR methods struggle to maintain their performance~\cite{dang2020,xu2022going}. Recently, wireless sensing methods~\cite{zhang2020human,yang2023sensefi,chen2018wifi,yang2022securesense} that leverage wireless signals, e.g., WiFi and radar, for human sensing, have drawn great attention as they are robust to illumination and are complementary to vision-based solutions. 

\begin{figure}[t]
  \centering
  \includegraphics[width=1.\linewidth]{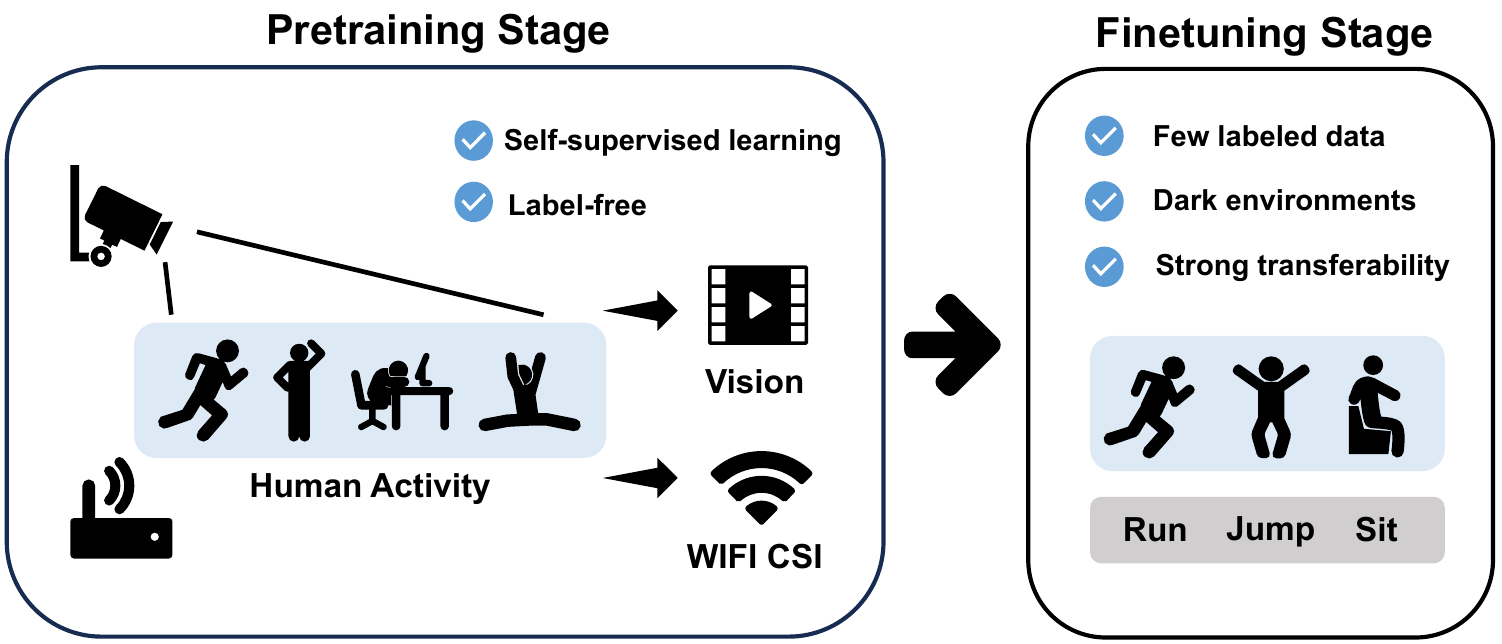}
  \caption{The application scenario of MI$^2$M framework. The WiFi and visual sensing data is trained in an unsupervised fashion, and the users only require few labeled data to finetune the HAR model. The proposed model can work well in dark environments with strong transferability.}
  \label{fig:0}
\end{figure}

To achieve better robustness in face of adverse conditions, WiFi-vision multimodal HAR system comes into existence~\cite{zou2019wifi}. When the vision modality is compromised by unfavorable environments, the WiFi modality can provide complementary sensing. Based on multimodal feature fusion, these approaches show better robustness and performance. However, they still face two major limitations. Firstly, different from computer vision field where many large-scale datasets are available, such datasets do not exist for WiFi that enable the models to generalize well across environments, and thus current WiFi sensing models are usually environment-dependent. Though some recent works aim at addressing this issue in WiFi sensing~\cite{wang2021multimodal,wang2022airfi}, they cannot be applied to multimodal HAR, i.e., the WiFi-vision HAR model. Secondly, existing works on WiFi-vision HAR systems only achieve simple feature-level concatenation~\cite{deng2022gaitfi} or decision-level fusion~\cite{zou2019wifi}. How to obtain a robust WiFi-vision representation by deeply exploring the cross-modal correlation remains a problem. 

To overcome the aforementioned challenges, we resort to a new way of unsupervised learning approach, which has been successfully utilized in vision-language models~\cite{huang2020pixel}. In our scenario, while labeled WiFi-vision data is cumbersome and costly to collect, unlabeled WiFi-vision data is easy to obtain. Recording the environment with human subjects is sufficient to obtain massive unlabeled WiFi-vision data without any human labor. To this end, the objective of this paper is to design an unsupervised learning framework that uses these unlabeled data to learn a WiFi-vision representation and capture the environmental dependency. With the pretrained model, users only need to calibrate several samples, and the whole WiFi-vision HAR system can work well, as shown in Fig.~\ref{fig:0}. Existing works on multimodal pretraining mainly deal with the connections between vision and language, yet WiFi-vision is more challenging without a deep understanding of WiFi modality. Hence, it is a non-trivial task to develop the first unsupervised WiFi-vision HAR model.

In this paper, we propose a new self-supervised learning framework for HAR, MaskFi, with a new data modeling approach namely \textbf{M}asked w\textbf{I}fi-v\textbf{I}sion \textbf{M}odeling (MI$^2$M). The proposed method learns the WiFi and video representation jointly without any annotation. Specifically, we first design an encoder, i.e., a neural network, to extract features from video and WiFi data, and tokenize the features into a joint feature representation. Then we propose to mask parts of the joint feature in a random fashion and train an encoder-decoder network to reconstruct the masked representation into the original representation. In this way, the encoder and the decoder capture the inner relationships between the WiFi and the video modalities. As the joint representation entangles two modalities, cross-modal learning is achieved in the proposed MaskFi. Furthermore, the designs of encoder and decoder in MaskFi are tailored by transformers that have a stronger capacity to learn cross-modal features through the self-attention mechanism. In practice, such unsupervised learning can be conducted automatically using the unlabeled data on the user side, and then the HAR model is finetuned using only several labeled samples to achieve satisfactory results. 

The main contributions of this paper are as follows:
\begin{itemize}
    \item We propose a self-supervised HAR system based on a novel multimodal pretraining strategy, MI$^2$M. As far as we know, MI$^2$M is the first model that incorporates self-supervised learning in WiFi-vision HAR systems and achieves the labor-efficient multimodal HAR task.

    \item The MI$^2$M is designed to learn the multimodal representation by reconstructing the masked multimodal representations. To complete the reconstruction, the encoder and decoder must capture the multimodal information and their cross-modal correlation at each frame. Without any labels, the encoder manages to restore more useful features that facilitate downstream HAR tasks.

    \item Extensive experiments are conducted in two real-world settings, and empirical results show that our method outperforms existing HAR methods based on WiFi and cameras, achieving state-of-the-art performance. 
\end{itemize}

\section{Related Work}
\subsection{Vision-based HAR Systems}
The most popular method in device-free human activity recognition is based on video captured by cameras. With the development of artificial intelligence algorithms, convolutional neural networks (CNNs) have gradually become the mainstream approach for visual-based human activity recognition~\cite{Beddiar2020}. Karpathy et al.~\cite{karpathy20} discovered that CNNs can automatically extract spatial information from video sequences, surpassing manually extracted features from individual frames. However, their approach focused only on single-frame information, lacking the extraction of temporal features. To address this, an improved network called C3D was introduced, which builds upon CNN to capture both spatial and temporal features from consecutive frames. Nevertheless, CNN with convolutional operators is limited by their receptive field. Building on this, Hussain et al.~\cite{hussain2022} proposed the use of transformer-based methods without convolutional operators to address this limitation. They utilized a pretrained vision transformer to extract both spatial and temporal features from video sequences and incorporated LSTM networks to capture long-range dependencies of actions. The main challenges faced by visual-based methods in HAR are recognition in low-light conditions and in the presence of occlusion. Introducing additional modalities in visual processing can effectively resolve this issue.

\subsection{WiFi-based HAR Systems}
With the ubiquity of network infrastructure, WiFi has been recognized as the most common wireless signal indoors. Compared to wearable devices~\cite{luo2021binarized,huang2022channel}, WiFi-based HAR solutions are convenient to use, non-intrusive, and more accurate due to more views of data. There are three types of information that can be used for HAR in WiFi: received signal strength indicator (RSSI), channel state information (CSI), and specialized radio hardware-based signals. RSSI has limited resolution, making it challenging to achieve fine-grained human action recognition. Specialized radio hardware-based signals require expensive installation and deployment of dedicated devices and are not commonly available in commercial WiFi deployments~\cite{yan2019}. CSI-based solutions~\cite{zhang2020human,yang2023sensefi,chen2018wifi,yang2022securesense} have been exploited to serve as a ubiquitous HAR system in the future.

Zhang et al.~\cite{zhang2020data} expanded a single WiFi CSI dataset into more activity data using eight data augmentation methods and utilized long short-term memory (LSTM) networks for classification to avoid overfitting issues commonly observed in small dataset training. Li et al.~\cite{li2019wi} proposed the WiFi motion model, which uses both phase and amplitude information from CSI to classify human actions. Different networks are employed to classify the two types of data, and the outputs of the classifiers are combined using a posterior probability-based combination strategy. Yang et al.~\cite{yang2022securesense} proposed adversarial augmentations to enhance the security of CSI-based HAR. Yadav et al.~\cite{yadav2022} introduced the CSITIME framework, considering the WiFi CSI activity recognition problem as a multivariate time series classification problem. Ji et al.~\cite{ji2022sifall} proposed a falling activity recognition system based on anomaly detection network. Zhang et al.~\cite{zhang2023location} proposed a location-independent HAR system based on graph neural networks. Whereas, the main challenge of using WiFi CSI signals for human activity recognition lies in the fact that CSI signals are not as intuitive as visual signals. Therefore, collecting and annotating data becomes more difficult Traditional data augmentation methods can help but they cannot simulate complex environmental conditions and a variety of actions, making the supervised learning-based models harder to generalize.

\subsection{Unsupervised Learning for Data-Efficient HAR}
Unsupervised learning aims to address the challenge of data annotation by leveraging the inherent characteristics of data to perform self-training without labels~\cite{soloe2020,jain2022}. Aditya Ramesh pointed out that the foundation of classification, recognition, and semantic segmentation problems lies in extracting structural and representative features by disregarding shallow details from the data~\cite{zhang2020data}. Saeed et al.~\cite{saeed2019} a simple binary classification task as an auxiliary task to supervise the self-supervised learning process.
Khaertdinov et al.~\cite{kha2021} proposed the contrastive self-supervised learning approach to sensor-based Human Activity Recognition for pretraining. They first applied random augmentation to the input data and then used a backbone network consisting of convolutional neural networks and transformer networks to match the augmented and original versions of the same input data. In WiFi-based HAR, the AutoFi~\cite{yang2022autofi} is firstly proposed self-supervised learning method that leverages mutual information and geometric structures for feature learning. While these works are proposed solely for vision or WiFi-based HAR, they cannot be directly applied to WiFi-vision HAR due to their multimodal nature, which motivates this work to explore the new era.

\begin{figure*}[ht]
  \centering
  \includegraphics[width=1.\linewidth]{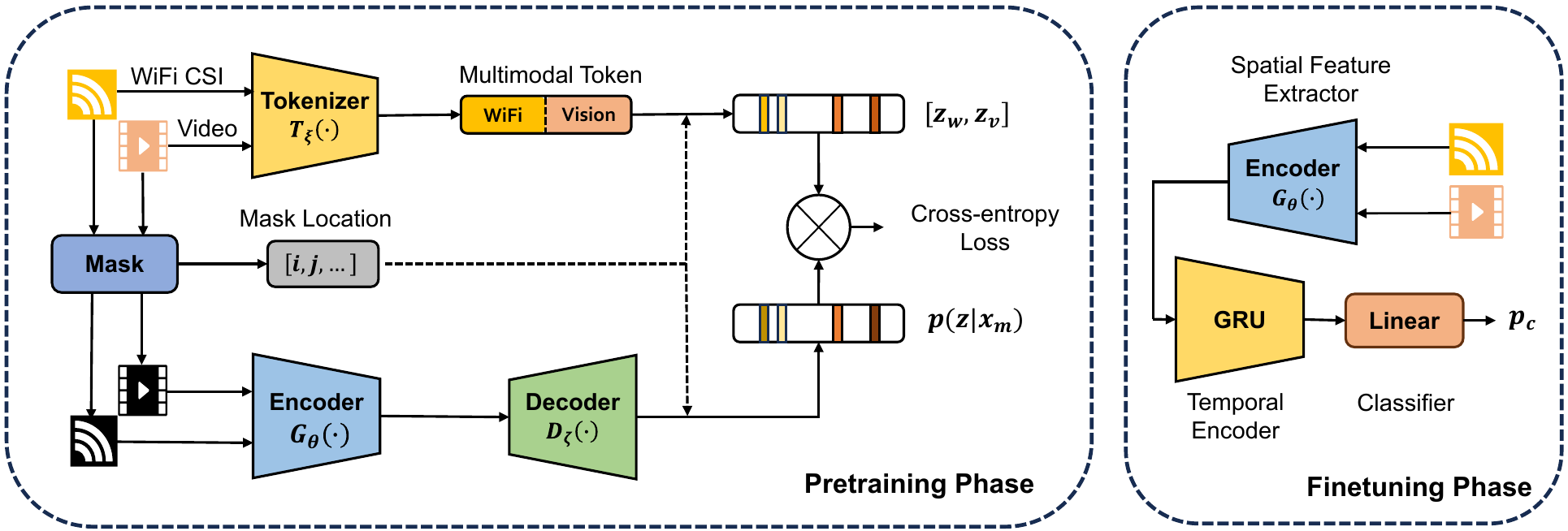}
  \caption{The illustration of the proposed MI$^2$M learning method.}
  \label{fig:framework}
\end{figure*}

\section{Method}
\subsection{Problem Formulation}
The objective of our work is to build a WiFi-vision HAR model with massive unlabeled data and a few labeled data. Therefore, as shown in Fig.~\ref{fig:framework}, the MaskFi consists of two parts: the pretraining phase (MI$^2$M) based on self-supervised learning and the finetuning phase for few-shot calibration. Let $X_w\in \mathbb{R}^{N_T \times N_s \times N_a}$ denote the WiFi CSI data where $N_T$ denotes the number of frames, $N_s$ denotes the number of subcarriers for a pair of antennas and $N_a$ denotes the number of antenna pairs. Let $X_v\in \mathbb{R}^{N_T \times H \times W}$ denote the video data where $H$ and $W$ denote the height and width of the video, respectively. 
In the pretraining phase, we first input the raw CSI data $X_w$ and the video signal $X_v$ into the tokenizer $T_\xi(\cdot)$ parameterized by $\xi$ to obtain their encoded representations $z_w$ and $z_v$, respectively. Then we mask the raw data and input it into a feature extractor $G_\theta$, parameterized by $\theta$, to obtain the features of the masked WiFi and image data at each frame. Then a decoder $D_\zeta$ parameterized by $\zeta$ is learned to map the masked feature into the original data that should be as same as the encoded representations $z_w$ and $z_v$. In the finetuning phase, as the first phase has trained the encoder to extract discriminative features at each frame, we simply use several samples collected by users to train a GRU for temporal feature extraction and a linear classifier after the pretrained encoder $G_\theta$. 

\subsection{Building WiFi and Vision Tokens}
To perform masked data prediction, we first design the model to learn sequential discrete tokens for the WiFi CSI and the image modality. This is achieved by a multimodal tokenizer $T_\xi$. As the tokenizer has been well studied in the computer vision field, we design the multimodal tokenizer according to the discrete variational autoencoder (dVAE) used in text-to-image generation~\cite{ramesh2021}. Specifically, we tokenize the WiFi CSI $X_w$ into $z_w=[z_w^1,z_w^2,...,z_w^N]\in \mathcal{V}^{N_s}$ where the vocabulary $\mathcal{V}^{N_s}$, \textit{i.e.,} the CSI codebook, is the set of discrete token indices. We design a multimodal tokenizer that maps each subcarrier value $x_w$ to a discrete token $z_w$ according to the CSI codebook. Then a decoder $p(x_w|z_w)$ learns to reconstruct the input subcarrier $x_w$ according to the token $z_w$. The learning objective of the tokenization process is written as:
\begin{equation}
    \min_\xi \mathbb{E}[\log p(x_w|z_w)].
\end{equation}
To enable the model training with discrete tokens, the Gumbel-softmax relaxation is employed~\cite{jang2016categorical}. The tokenizer and the decoder all consist of two convolutional layers and are easy to train. The same tokenization process applies to the vision modality so that the image $x_v$ is tokenized into a discrete token $z_v$. 

\subsection{Spatial Encoding of WiFi and Vision}
Though the tokenizer can be applied to the whole data frame, it loses the spatial feature during pretraining. Therefore, we propose to divide the CSI and image data into patches. When our model can reconstruct a patch according to its neighbor patches, it has the capacity to capture the spatial features within the subcarrier and the image pixel dimension. To this end, we split each CSI into $P_w\times P_w$ patches and each image into $P_v\times P_v$ patches. The tokenizer is applied to all patches so that they can be masked and predicted by the encoder in our pretraining.

\subsection{Multimodal Pretraining for WiFi-Vision HAR}
To model the multimodal features of WiFi CSI and vision image data, we propose the \textit{Masked wIfi-vIsion Modeling} (MI$^2$M) task. The objective is to train the encoder $G_\theta$ that is able to extract features from two modalities in an unsupervised manner. As demonstrated in Fig.~\ref{fig:framework}, for a sample at one frame, we divide the WiFi CSI and vision image into $N_w$ CSI patches and $N_v$ image patches, respectively. These patches are then tokenized into CSI tokens $\{z^i_w\}^{N_w}_{i=1}$ and visual tokens $\{z^i_v\}^{N_v}_{i=1}$. We randomly mask the $\alpha$ proportion of both patches and obtain the masked positions as $\mathcal{M}_w$ and $\mathcal{M}_v$. To encourage the encoder to restore these masked patches, we replace these masked positions with learnable embeddings $\{m^i\}^{\alpha N_w N_v}_{i=1}$, obtaining the corrupted multimodal sample $x_m=\{x_w\}^{N_w}_{i=1} \cup \{x_v\}^{N_v}_{i=1} \cup \{m^i\}^{\alpha (N_w+N_v)}_{i=1}$. Then these patches are fed into the encoder, which is a transformer-based network introduced later. For each frame, both WiFi CSI and image patch are fed into the encoder together, and thus the transformer can learn how to model both modalities together by restoring the masked positions. The output of the encoder is the final hidden vectors $\{h^i\}^{N_w+N_v}_{i=1}$. As we expect the masked positions can be reconstructed, the decoder network is a simple softmax classifier to predict the corresponding tokens, which is formulated as
\begin{equation}
    p(z|x_m)=\sigma({W_d}h+b_d),
\end{equation}
where $\sigma$ is the softmax function, and $W_d,b_d$ are the parameters of the classifier. The learning objective is to maximize the log-likelihood of the correct WiFi and visual tokens given the masked multimodal data:
\begin{equation}\label{eq:mim}
    \max_{\theta,\zeta} \mathbb{E}\bigg[\sum_{i\in \mathcal{M}_w \cup \mathcal{M}_v} \log(z_i|x_m) \bigg].
\end{equation}
The design of MI$^2$M is inspired by the famous masked language modeling in BERT~\cite{devlin2018bert} and masked image modeling in BEIT~\cite{beit2021}. The difference is that we manage to train the encoder on multimodal data. The model is able to learn the relationships between two modalities by predicting the masked positions of WiFi and vision modalities together. Apart from the masked modeling, the transformer network has the advantage of capturing cross-modal features and accepting multimodal inputs, so we choose it as the encoder in our model.

We briefly introduce the encoder architecture composed of $L$ layers of standard transformer blocks~\cite{vas2017} that enable self-attention mechanisms. A stack of multi-head self-attention and linear layers is followed by layer normalization to avoid possible gradient explosion or gradient vanishing problems. The training parameters of these transformer blocks are three matrices $Q,K,V$, and the attention mechanism is achieved by:
\begin{equation}
S=\sigma(\frac{QK}{\sqrt{d_k}})V.
\end{equation}
We use multi-head attention to further enhance the capacity to capture cross-modal features. The details of the transformer network can be referred to~\cite{vas2017}.

\subsection{Finetuning A Classifier By Users}

After the MI$^2$M is completed, we require the user to label few data for training. However, the proposed MI$^2$M only extracts robust multimodal features from one frame of two modalities at the same timestamp, and therefore cannot capture temporal information. The temporal features across frames reflect the sequence of actions and are of paramount importance to WiFi-vision human activity or gesture recognition. To this end, we consider adding a temporal feature extractor after the transformer-based encoder. Though Long Short-Term Memory (LSTM) is widely used in time-series analysis, it causes large computational complexity. We choose to leverage Gate Recurrent Unit network (GRU) which only consists of one gate to forget and select memory simultaneously.

Denote the multimodal features of consecutive frames as $f_m=\{f^1,f^2,...,f^N_t\}$. They are fed into the GRU computed as:
\begin{equation}
r_t=\delta(W_rf_t+U_rs_{t-1}+B_r),
\end{equation}
\begin{equation}
z_t=\delta(W_zf_t+U_zs_{t-1}+B_z),
\end{equation}
\begin{equation}
h_t=tanh(W_zf_t+U_h(r_t s_{t-1})+B_h),
\end{equation}
\begin{equation}
    s_t=(1-z_t) s_{t-1}+z_t h_t,
\end{equation}
where $W_r,W_z,U_r,U_z,B_r,B_z$ represent weights of GRU, and $s_{t}$ represents the states at the frame $t$. The final output of the GRU network is:
\begin{equation}
o_t=\delta(W_o s_t+B_o),
\end{equation}
where $\delta(\cdot)$ is a sigmoid function, and $W_o,B_o$ are the weights of the GRU. The output is fed into a linear layer followed by a softmax function, obtaining the prediction probability $p_c$. To optimize the GRU and the linear classifier, we use the cross-entropy loss between the prediction and the ground truth label $y_c$ formulated as:
\begin{equation}
	\mathcal{L}_{ce}=-\mathbb{E}_{y}\sum_k \big[ \mathbb{I}[y=k] \log(p_c) \big],
\end{equation}
where $\mathbb{I}[y=k]$ means a 0-1 function that outputs 1 for the correct category $k$. 

We summarize the whole algorithm of MI$^2$M and the finetuning in Algorithm~\ref{al:1}. In practice, the WiFi and camera sensors can capture massive unlabeled data in an unknown environment, which enables MI$^2$M pretraining. Then the user uses only few samples to train the GRU and classifier for customized action recognition.

\begin{algorithm}[ht]
\caption{Masked WiFi-Vision Modeling}\label{al:1}
\SetAlgoLined
\KwIn{Unlabeled WiFi data $X_w$ and visual data $X_v$}
\SetKwProg{Fn}{Function}{:}{end}
\Fn{Pretraining}{
  Initialize parameters $T_\xi,G_\theta,D_\zeta$\;
  Divide the data into patches\;
  Tokenize the multimodal data into $z_w,z_v$\;
  \For{$epoch$}{
    Randomly choose the masked positions $\mathcal{M}_w,\mathcal{M}_v$\;
    Obtain the corrupted sample $x_m$\;
    Inference the encoder and obtain the prediction $p(z|x_m)$\;
    Maximize the likelihood in Eq.(\ref{eq:mim})\;
  }
}
\SetKwProg{Fn}{Function}{:}{end}
\KwIn{Labeled data $X_v,X_w,y_c$, pretrained parameters $G_\theta$}
\Fn{Finetuning}{
  Initialize parameters of GRU and classifier\;
  \For{$epoch$}{
    Obtain the multimodal features $f_m$\;
    Obtain the prediction $p_c$\;
    Calculate the loss $\mathcal{L}_{ce}$\;
    Update parameters of GRU and classifier\;
  }
}
\end{algorithm}

\section{Experiments}
\subsection{Setup}
\subsubsection{Datasets and Scenario}
\begin{figure}[t]
  \centering
  \includegraphics[width=0.45\textwidth]{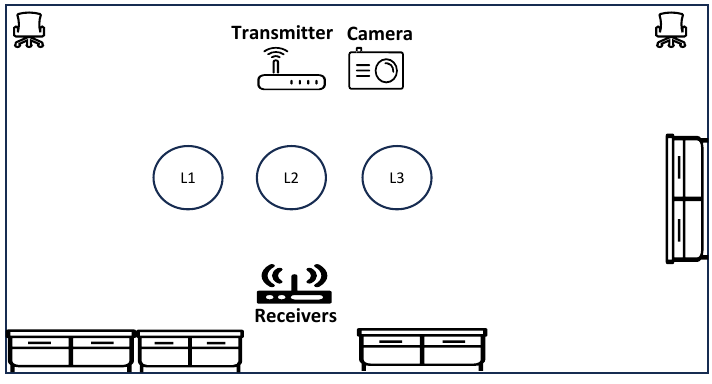}
  \caption{The layout and settings of WV-Lab Dataset.}
  \label{fig:layout}
\end{figure}

\textbf{WV-Lab Dataset.}
We evaluate our method using a self-collected dataset, WV-Lab, which is collected in laboratory environments. The system consists of a WiFi CSI system~\cite{yang2018device} based on Atheros CSI tool~\cite{xie2015precise} and an Intel RealSense D435. The WiFi CSI system is built on 2 TP-Link N750 WiFi access points (5GHz) with 3 receiver antennas and 1 transmitter antenna. With 80MHz bandwidth, we can collect 114 subcarriers per pair of antennas. It comprises 132 minutes of synchronized video and WiFi CSI data streams obtained from 22 volunteers in 2 different environmental settings. The layouts of the environment are shown in Fig.~\ref{fig:layout}. The different environments are made by changing the positions of the systems and moving the furniture. The 22 participants perform 6 daily actions, with each action spanning a duration of 1 minute. These actions encompass the left arm rotating clockwise and counterclockwise, the right arm rotating clockwise and counterclockwise, walking in place, and legs shaking. The video and WiFi CSI data streams are sampled with the rate of 100 Hz and 1000 Hz, respectively, and thus each frame of the video corresponds to ten packets of CSI data. 

\begin{figure*}[t]
  \centering
  \includegraphics[width=0.9\textwidth]{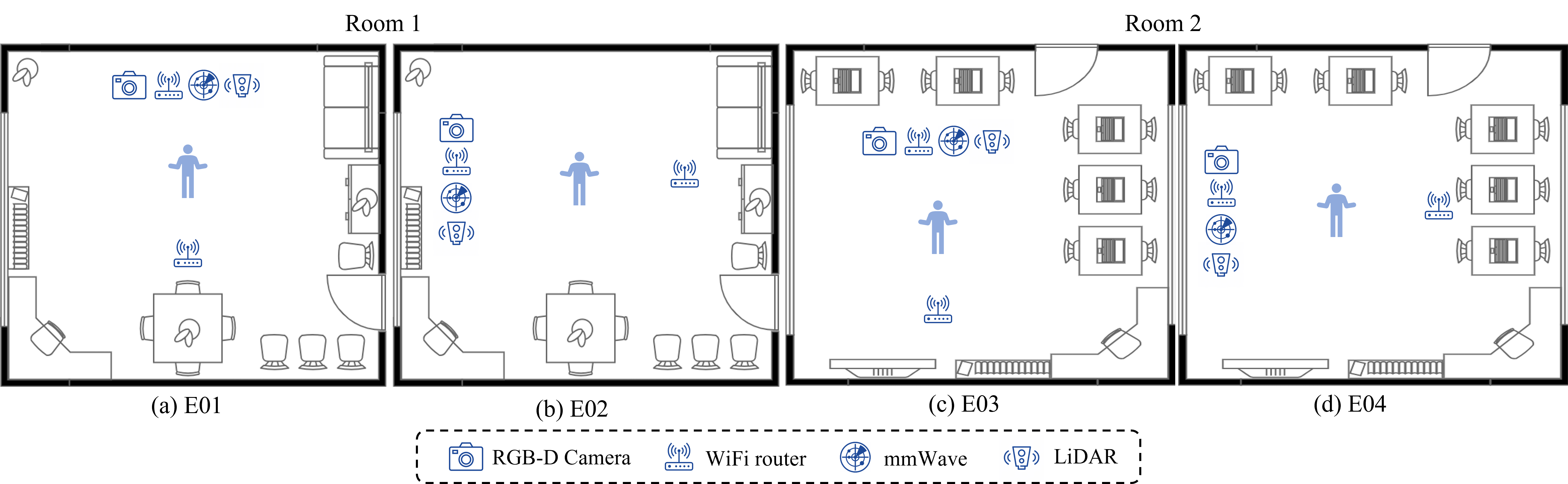}
  \caption{The layout and settings of MM-Fi Dataset~\cite{yang2023mmfi}.}
  \label{fig:mmfi-layout}
\end{figure*}

\noindent
\textbf{MM-Fi Dataset}~\cite{yang2023mmfi}.
To evaluate the effectiveness of our method, we also use a large-scale public WiFi-vision dataset, MM-Fi, which contains 40 volunteers performing 27 actions. The datasets are collected in four distinct environments, as shown in Fig.~\ref{fig:mmfi-layout}. We only use the WiFi and vision modality in MM-Fi. The action categories of MM-Fi Dataset include chest expanding horizontally, chest expanding vertically, left side twist, right side twist, raising left arm, raising right arm, waving left arm, waving right arm, picking up things, throwing toward left side, throwing toward right side, kicking toward left direction using right leg, kicking toward right direction using left leg, bowing, stretching and relaxing in free form, mark time, left upper limb extension, right upper limb extension, left front lunge, right front lunge, both upper limbs extension, squat, left side lunge, right side lunge, left limbs extension, right limbs extension and jumping up. The Dataset comprises a total of around 320,000 frames. MM-Fi is a challenging dataset due to the large number of activities and severe environmental differences.

\begin{table*}[t]
\centering
\caption{Class-wise accuracy (\%) comparison on WV-Lab Dataset.}\label{tab:wv-lab}
\begin{tabular}{c|c|c|cc|cc|cc|c}
\toprule
 & & & \multicolumn{2}{c|}{Left Arm Rotation} &\multicolumn{2}{c|}{Right Arm Rotation}&\multicolumn{2}{c|}{Leg Movements} \\
Method &Modality&Unsupervised & Clockwise&Counter&Clockwise&Counter & Shaking & Walking & Average \\
\midrule

ViT\cite{hussain2022vi}&Vision&$\times$	&96.74	&94.11	&93.65	&96.14	&96.92	&97.12 &95.78\\
3D-ResNet-18\cite{hara2017}&Vision&$\times$	&92.31	&93.24	&94.78	&94.66	&97.68	&96.55	&94.87\\
Darklight\cite{chen2021d}&Vision&$\times$	&97.46	&96.98	&95.62	&97.11	&98.62	&97.65	&97.24\\
Pro-CNN\cite{moshiri2021}&WiFi&$\times$	&91.41	&92.16	&93.33	&91.55	&97.50	&93.37	&93.22 \\
THAT\cite{li2021two}&WiFi&$\times$	&99.47	&99.48	&98.32	&92.59	&99.69	&96.35	&97.70 \\
ID-CNN\cite{wang2019joint}&WiFi&$\times$	&91.93	&96.25	&97.45	&98.76	&90.92	&83.23	&93.85 \\
MaskFi &WiFi+Vision&$\checkmark$&94.69	&97.37	&97.13	&97.92	&99.89	&98.66	&97.61\\

\bottomrule
\end{tabular}
\end{table*}

\subsubsection{Evaluation Protocol}
We adopt the true positive rate, \textit{i.e.,} the accuracy, as the evaluation criterion to assess the recognition performance of the MaskFi framework. The true positive rate represents the proportion of correctly classified samples among all samples. The mean accuracy across three runs is reported. We use 80\% of the frames in each data for MI$^2$M pretraining. The left 20\% frames serve as the testing data. In the finetuning, we only feed 1 minute of data per class. Action samples for testing are obtained by video segmentation every 8 frames. To demonstrate that our method can facilitate various downstream tasks, we conducted two recognition tasks: (1) human activity recognition and (2) simultaneous activity recognition and human identification.

\subsubsection{Implementation Details}
Here, we provide the experimental details of the MaskFi framework. The input size for video frames is $3\times224\times224$, while the input size for WiFi CSI frames is $3\times114\times10$. The codebook size is set to 8192. During the pretraining phase, we masked $\alpha=40\%$ of the data. The learning rate was set to 5e-4, the batch size is set to 128, and a total of 80 epochs are trained on a workstation with 4 NVIDIA RTX3090 GRUs. In the subsequent finetuning phase, the sequence length in the GRU is set to 8, and the batch size is set to 32. With a learning rate of 4e-4, we find that the convergence can be achieved using only 10 epochs since the pretraining phase has already been completed. We employ the Adam optimizer during the pretraining and finetuning.

\subsubsection{Baselines}
To validate the effectiveness of the MaskFi framework and the pretraining phase, we compared our approach with various existing methods for human activity recognition based on WiFi CSI and video. The vision-based baselines include Vision Transformer (ViT)~\cite{hussain2022vi}, 3D Residual Network (3D-ResNet-18)~\cite{hara2017}, and Darklight~\cite{chen2021deep} (specifically designed for dark environments). The WiFi-based baselines include CNN-based solutions (the Pro-CNN network~\cite{moshiri2021} and ID-CNN network~\cite{wang2019joint}) and transformer-based solution (THAT)~\cite{li2021two}. As there does not exist any unsupervised method for WiFi-vision HAR, these methods are all supervised learning method trained with all training data (used as unlabeled data in our method).

\begin{figure}[t]
  \centering
  \includegraphics[width=0.4\textwidth]{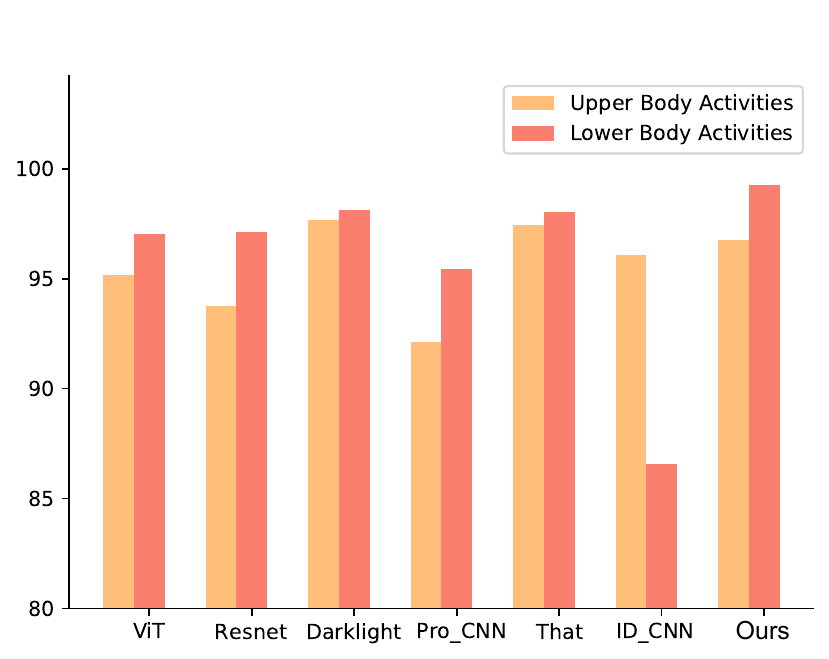}
  \caption{Accuracy (\%) comparison of two categories on WV-Lab Dataset.}
  \label{fig:wv-lab-category}
\end{figure}

\subsection{Evaluation on WV-Lab}
We divide the six human activities in WV-Lab into three categories: leg movements, left-arm movements, and right-arm movements, and we report the class-level and average accuracy. The experimental results and comparison are presented in Table~\ref{tab:wv-lab} and Fig.~\ref{fig:wv-lab-category}. It is observed that our approach achieves a very competitive performance on average. Specifically, most methods can distinguish leg swinging and walking actions well due to their significant differences. However, in arm movements, vision-based network methods outperform WiFi-based methods, since the CSI data is not as high-resolution as videos. Whereas, our proposed MaskFi framework absorbs the advantages of both modalities and shows strong recognition capacity for both arm and leg movements. It achieves an average accuracy of 97.61\% with only a small amount of labeled data for finetuning.

\begin{table*}[!t]
\centering
\caption{Accuracy (\%) comparison on MM-Fi Dataset}\label{tab:mm-fi}
\begin{tabular}{c|c|c|c|c}
\toprule
Method &Modality&Unsupervised& Activity Recognition & Activity \& Subject Identification  \\
\midrule
ViT~\cite{hussain2022vi}	&Vision&$\times$&95.07	&91.58  \\
3D-ResNet-18~\cite{hara2017}&Vision&$\times$	&94.56	&95.78  \\
Darklight~\cite{chen2021d}&Vision&$\times$	&96.85	&95.23  \\
Pro-CNN~\cite{moshiri2021}&WiFi&$\times$	&93.17	&94.12  \\
THAT~\cite{li2021two}&WiFi&$\times$	&97.21	&96.25  \\
ID-CNN~\cite{wang2019joint}&WiFi&$\times$	&92.03	&95.29  \\
MaskFi &WiFi+Vision&$\checkmark$&96.82	&97.27\\
\bottomrule
\end{tabular}
\end{table*}

\subsection{Evaluation on MM-Fi}
MM-Fi Dataset contains more 40 challenging human activities where the subject moves the arm, leg, and body together. As there are 40 subjects involved, we also conduct simultaneous activity recognition and human identification which is to identify the category of the activity and who performs this activity. As shown in Table~\ref{tab:mm-fi}, the overall performances of all methods are lower than those of WV-Lab due to more difficult categories. The proposed method achieves an average accuracy of 96.82\% for activity recognition, even outperforming many supervised approach using vision or WiFi, which demonstrates that our approach can handle more difficult tasks with strong capacity of learning spatial and temporal features. It is surprising that our method even achieve a better performance of 97.27\% on simultaneous activity and human identification task. The reason might be the more abundant annotations with both subject and activity labels that better guide the model training. This justifies that the representation learned by MI$^2$M can enable more downstream tasks apart from single activity recognition.

We also notice that WiFi-based subject identification methods achieve higher accuracy compared to vision-based approaches. We observe the data and find that the faces of subjects in MM-Fi dataset are masked, making it difficult to discern fine facial features. As a result, body size, clothing and posture habits become the primary distinguishing factors among different volunteers. The WiFi CSI can capture more movements instead of the appearance captured by camera, which explains why WiFi-based method can better perform subject identification. Our method leverages both WiFi and camera, achieving the state-of-the-art accuracy when it is compared to other WiFi-based and vision-based methods.

\subsection{Analytics of Dark Environmental Results}
We conduct experiments in dark environments to demonstrate the effectiveness of our method in adverse conditions. To this end, we evaluate our method on self-collected dark environments in WV-Lab, and apply the gamma enhancement for MM-Fi to obtain a dark version of MM-Fi (activity recognition task). The results have been shown in Table~\ref{tab:dark}. It is shown that our method can still achieve the accuracy of 92.17\% and 90.43\% on WV-Lab and MM-Fi, respectively. Though the performances are slightly lower than those of normal illumination conditions, the results are still acceptable. The WiFi CSI that is not affected by lighting conditions helps compensate for the video modality. Therefore, the MaskFi framework, which incorporates multimodal fusion, exhibits stronger robustness in adverse conditions compared to a single-modality network.

\begin{table}[!ht]
\centering
\caption{Assessment under varying lighting conditions.}\label{tab:dark}
\begin{tabular}{cc|c}
\toprule
Pretrain Dataset & Environment  & Accuracy (\%) \\
\midrule
WV-Lab Dataset  & Normal condition & 97.61  \\
WV-Lab Dataset  & Dark condition   & 92.17  \\
MM-Fi Dataset & Normal condition  & 96.82  \\
MM-Fi Dataset & Dark condition  & 90.43  \\
\bottomrule
\end{tabular}
\end{table}

\subsection{Analytics of Cross-Environment Evaluation.}
The design of MaskFi is to enable the system without massive annotations in a new environment. In previous studies, we assume the massive unlabeled data comes from the new environment so that our method can learn the environmental information, which has been demonstrated to be very effective. Here we consider another situation: is it possible to train the model in a large public dataset and then finetune the model in a new environment with few data? If this can be achieved, then users in practice do not need a long period to set up the environment. To this end, we conduct the cross-environment evaluation: training the model in one dataset, and then finetuning and evaluating the model in a new environment. We still use WV-Lab and MM-Fi. As shown in Table~\ref{tab:environment}, when the model is pretrained on the large dataset MM-Fi, it shows 95.87\% accuracy on the new environment WV-Lab after finetuning. Compared to the vanilla settings with 97.61\% accuracy, the performance only drops by 1.74\%, demonstrating the effectiveness of our MI$^2$M in cross-environment scenarios. In the opposite, when the pretraining dataset is the WV-Lab, the model finetuned on MM-Fi achieves 93.15\% with a accuracy drop of 3.67\%. This indicates that the pretraining should be conducted on a larger dataset to have better model transferability.

\begin{table}[!ht]
\centering
\caption{Evaluation across different datasets.}\label{tab:environment}
\begin{tabular}{cc|c}
\toprule
Pretrain Data & Finetune Data  & Accuracy (\%) \\
\midrule

WV-Lab  & WV-Lab  & 97.61  \\
WV-Lab  & MM-Fi & 93.15  \\
MM-Fi & WV-Lab  & 95.87  \\
MM-Fi & MM-Fi & 96.82  \\
\bottomrule
\end{tabular}
\end{table}



\section{Conclusion}

This paper proposes the MaskFi framework, a novel unsupervised learning framework for multimodal human activity recognition with both WiFi and vision modalities. To this end, the two modalities of data are tokenized into a multimodal representation. Then we mask the multimodal data and feed the masked data into a transformer-based network. By learning how to reconstruct the masked multimodal data into the original multimodal representation, the encoder can capture the multimodal correlation and features. Then in the finetuning phase, a temporal feature extractor and a simple classifier are trained with less data. The experiments are conducted in the real world, and extensive results on our data and public dataset show that our approach achieves 97.61\% accuracy on WV-Lab and 96.82\% accuracy on MM-Fi regarding the HAR task, which demonstrates the effectiveness of our framework in practice.

Possible future research directions include more downstream tasks besides human activity recognition, such as gesture recognition and pose estimation. This can further demonstrate the generalization ability of MI$^2$M. The application scenarios of MI$^2$M are also worthy of further research, such as expanding from single-person HAR to multi-person HAR. Moreover, whether our framework applies to more multimodal HAR with mmWave radar and LiDAR is also worth exploiting.

\bibliographystyle{plain}
\bibliography{reference}

\begin{IEEEbiography}[{\includegraphics[width=0.9in,clip,keepaspectratio]{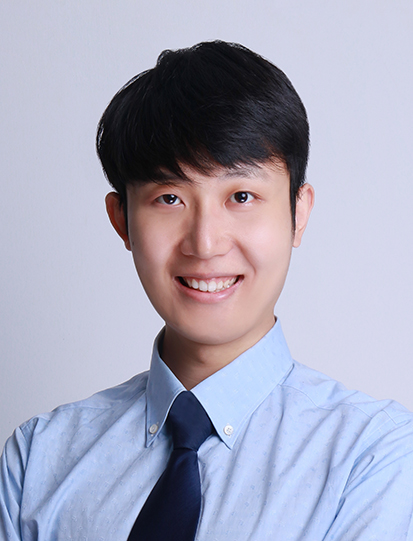}}]{Jianfei Yang}
	received the B.Eng. from the School of Data and Computer Science, Sun Yat-sen University in 2016, and the Ph.D. degree from Nanyang Technological University (NTU), Singapore in 2021. He received the best Ph.D. thesis award from NTU. He used to work as a senior research engineer at BEARS, the University of California, Berkeley. His research interests include deep transfer learning with applications in Internet of Things and computer vision. He won many International AI challenges in computer vision and interdisciplinary research fields. Currently, he is a Presidential Postdoctoral Research Fellow and an independent PI at NTU.
\end{IEEEbiography}

\begin{IEEEbiography}[{\includegraphics[width=0.9in,clip,keepaspectratio]{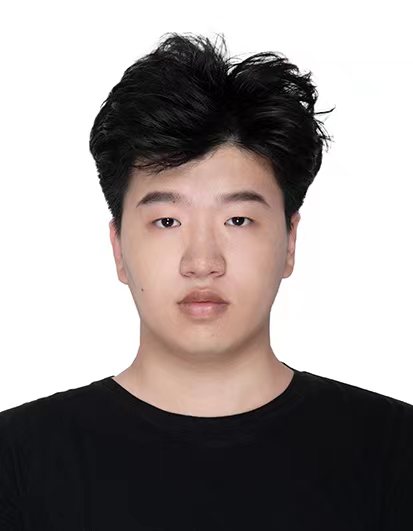}}]{Shijie Tang}
	received the B.Eng. from the School of Control Science and Engineering at Shandong University in 2022, and the Msc. from the School of Electrical and Electronic Engineering, Nanyang Technological University (NTU), Singapore in 2023. His research interests include deep learning, the Internet of Things, and computer vision.
\end{IEEEbiography}

\begin{IEEEbiography}[{\includegraphics[width=0.9in,clip,keepaspectratio]{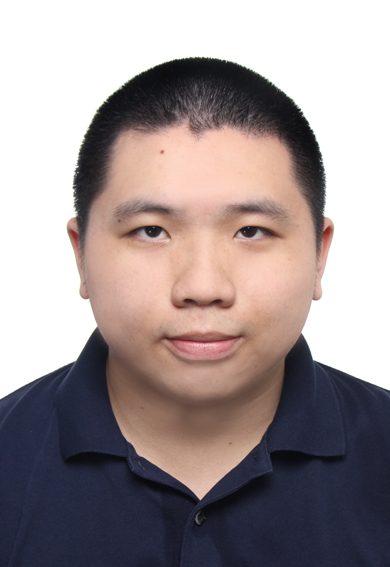}}]{Yuecong Xu}
    received the B.Eng. from the School of Electrical and Electronic Engineering, Nanyang Technological University, Singapore in 2017, and the Ph.D. degree from Nanyang Technological University (NTU), Singapore in 2021. He was the receiver of the Nanyang President's Graduate Scholarship. His research focuses on multi-modal video and time-series data analysis based on deep learning and transfer learning. He was the co-organizer of the UG2+ Challenge for Computational Photography and Visual Recognition, held in conjunction with CVPR 2021 and 2022. He used to work as a research scientist at the Institute for Infocomm Research. Currently, he is a lecturer at NTU.
\end{IEEEbiography}

\begin{IEEEbiography}[{\includegraphics[width=1in,height=1.25in,clip,keepaspectratio]{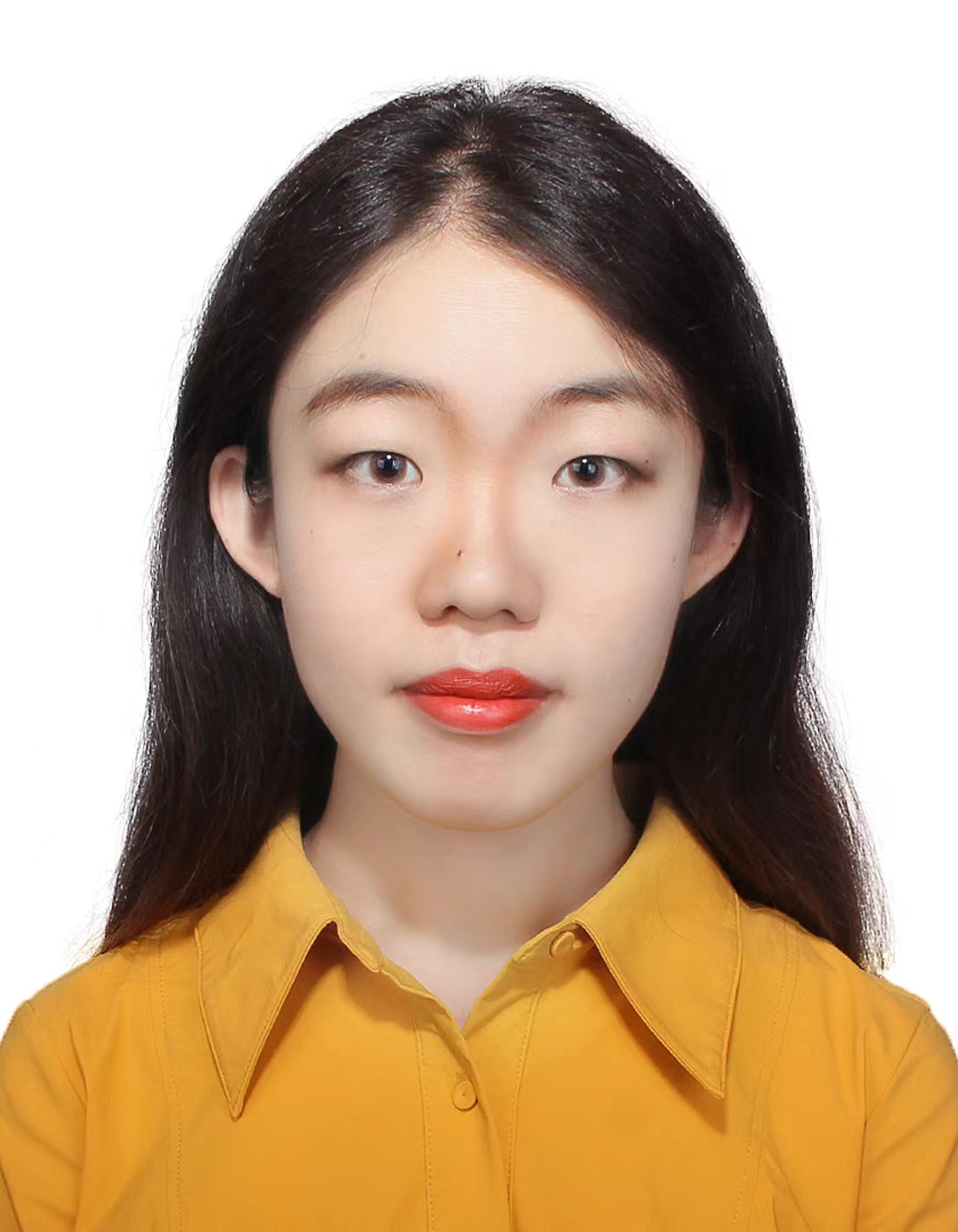}}]{Yunjiao Zhou}
	received the B.Eng. from the School of Information Science and Technology, Southwest Jiaotong in 2021, and the Msc. degree in Electrical and Electronic Engineering from Nanyang Technological University in 2022. She is currently working as a research associate in the Internet of Thing laboratory, Nanyang Technological University. Her research interests include multi-modal learning for multi-sensor applications and human perception in smart home.
\end{IEEEbiography}

\begin{IEEEbiography}[{\includegraphics[width=1in,height=1.25in,clip,keepaspectratio]{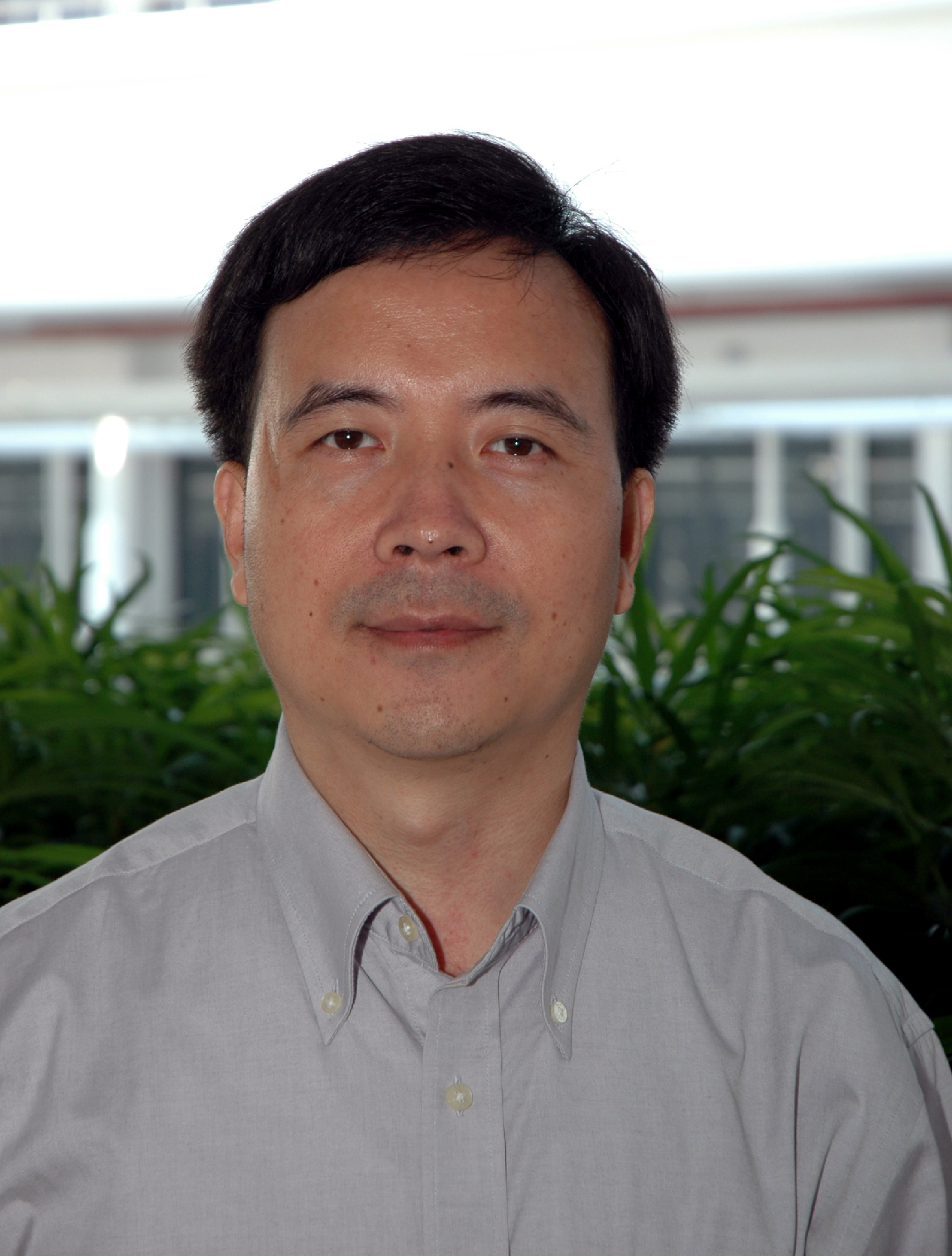}}]{Lihua Xie}
	received the B.E. and M.E. degrees in electrical engineering from Nanjing University of Science and Technology in 1983 and 1986, respectively, and the Ph.D. degree in electrical engineering from the University of Newcastle, Australia, in 1992. Since 1992, he has been with the School of Electrical and Electronic Engineering, Nanyang Technological University, Singapore, where he is currently a professor and served as the Head of Division of Control and Instrumentation from July 2011 to June 2014. He held teaching appointments in the Department of Automatic Control, Nanjing University of Science and Technology from 1986 to 1989 and Changjiang Visiting Professorship with South China University of  Technology from 2006 to 2011.
	
	Dr Xie's research interests include robust control and estimation, networked control systems, multi-agent control and unmanned systems. He has served as an editor of IET Book Series in Control and an Associate Editor of a number of journals including IEEE Transactions on Automatic Control, Automatica, IEEE Transactions on Control Systems Technology, and IEEE Transactions on Circuits and Systems-II. Dr Xie is a Fellow of IEEE and Fellow of IFAC.
\end{IEEEbiography}

\end{document}